Regularity of Position Sequences 1 of 12

# Regularity of Position Sequences

Manfred Harringer (*)

**Abstract**

A person is given a numbered sequence of positions on a sheet of paper. The person is asked, “Which will be the next (or the next after that) position?” Everyone has an opinion as to how he or she would proceed. There are regular sequences for which there is general agreement on how to continue. However, there are less regular sequences for which this assessment is less certain. There are sequences for which every continuation is perceived to be arbitrary. I would like to present a mathematical model that reflects these opinions and perceptions with the aid of a valuation function. It is necessary to apply a rich set of invariant features of position sequences to ensure the quality of this model. All other properties of the model are arbitrary.

## 1. Introduction

What is a pattern; what is a structure; what is regularity? There are at least partial answers to such questions. We know that snow has a structure because each snowflake is composed of regular hexagonal ice crystals. The bit sequence (1,1,1,1,1,1,1,1) is regular because it can be described thus: it consists solely of ones, a total of eight. The sequence of real numbers (2,2,3,3,4,4,5,5) demonstrates an easily recognisable regularity, on the basis of which one could continue with (...,6,6,7,7,...).

Mathematical concepts have been used to approach these issues. In the case of the snow flake it is symmetry. A regular hexagon has a symmetry group which transfers this pattern into itself. With the bit sequence we are reminded of the concept of Kolmogoroff complexity; the above bit sequence can be described “in a few words”. There is an extensive theory which deals with the problem of continuation of number sequences (see [2], [9]), from regressions (for example, Fourier transformation) to considerations of entropy ([4] S, 92 ff.).

On the other hand, the recognition of patterns, structures and uniformities is a typical phenomenon of human intelligence. In this respect something has been lost in the above mentioned mathematical concepts. In this form the concept of symmetry is too one-sided and rigid. The other ideas (complexity concepts, regression processes) reach far beyond that which can be achieved with normal human intelligence (without computers). I will try to demonstrate a closer relationship. For that reason I will limit myself to a very particular phenomenon of human intelligence.

A person is given a numbered sequence of positions on a sheet of paper and is asked which is the next (or next after that) position. Every person has an opinion as to how he or she should continue. Similar tasks are sometimes a part of intelligence tests (example in [5]). The goal is to find a model that describes this opinion. For this purpose a certain amount of idealisation is necessary. The positions are represented as fields on an n*n chessboard; n = 16 is a good start. If they are arranged in customary chess notation, A1,B2, C3, D4, E5, F6, it is agreed that the continuation will be G7,H8,I9…. Patterns or rules are recognisable which present a path to continuation. B2, C3, D2, E3 form a zigzag trail which can be continued with F2, G3. A further example is E6, D4, F3, H2, G4, F2, a sequence of moves with a knight which can be continued by a sequence of permissible knight moves, or G6, E3, H11, B4, L5, C9. These positions have been chosen more or less by chance. Thus no given continuation should present itself particularly strongly. The result of such deliberations concerning the continuation of position sequences must now be represented by a valuation function which determines the degree of suitability for each possible continuation; especially, the order of priority generated by the degree of suitability should correspond to personal preference.

(*)EMail manfred.harringer@t-online.de

I have developed a program system “EE” by which various versions of a valuation function can be implemented and tested. That led me to the following conclusion: It is decisive for the quality of a valuation function that it uses preferably those features of position sequences which are invariant under translation, rotation and stretching. A large supply of such operators is required. Each operator contributes to the total valuation of a potential continuation. All contributions are summarised by taking the mean value.

There is good reason to rely on invariant features. If I prefer a continuation of a sequence, I will also prefer the transformed continuation of the transformed sequence. The valuation function should reflect those invariants, although the restriction of sequences in the entire plane to those on a 16*16 chess board introduces a certain perturbation effect. It is easy to find such invariants for positions. For that reason it would seem wise to utilise such invariants in order to guarantee the invariance of the valuation function.

In todays physics symmetry considerations are essential. The continuation problem confirms the importance of symmetry considerations of this kind: The objects are not symmetric, but the relevant features are.

The solution of the problem using the program system “EE” has some restrictions. Here I am interested primarily in short sequences, that is, in sequences with a length of 4 to 8. In longer sequences other factors would certainly also be involved. I am restricted to geometric considerations. I do not take into consideration whether the coordinates of the positions may be prime numbers or something similar, i.e. mathematical concepts, as typical in [2]. I do not discuss special problems coming from the boundary of the chessboard or its granularity. It is sufficient to state, that some problems may be transformed by using a chessboard with higher granularity, with positions in the middle of the chessboard.

It is difficult to compare my results with others. There is few theoretical background for my feature selection process available. Practical background is not available, too. With simple examples there is no chance to compare different prediction results. For less simple examples there is no objective measure to compare the quality of different results, and it is an own problem to find such measures by psycological researches. May be, it is not reasonable to search for such measures.

In part 2. I would like to give some examples of how far we can go with the valuation function. In part 3. I describe the feature selection and the construction of the valuation function by these features. This construction may be replaced by many other methods. In part 4. I present advanced results. I mention numerical results about the importance of invariant features. I show, how far the concepts of a stage is learned, and I try to show the path to random sequences. In Part 5. I try to account some links to other researches.

## 2. Examples

Given a numbered position sequence $c = (c_0, \ldots, c_{n-1})$, a valuation function for arbitrary sequences is calculated by the program "EE". The diagrams contain the sequence, which is the base of the valuation and the sequence, which is the base of continuations, with position colors from dark red to light red. The values of all possible continuations are marked up by color blue, the positions with top values are enumerated, and the distribution of the values of all 256 possible continuation positions is shown.

valuation by

```
[ 0] ( 4, 4)
[ 1] ( 6, 4)
[ 2] (11, 4)
[ 3] (11, 6)
[ 4] (11,11)
```

sequence

```
[ 0] ( 4, 4)
[ 1] ( 6, 4)
[ 2] (11, 4)
[ 3] (11, 6)
[ 4] (11,11)
```

top values 1. - 5.

```
( 9,11):  2.1427
(13,11):  0.4973
( 8,11):  0.4223
(10,11):  0.2066
( 9,12):  0.0530
```

positions

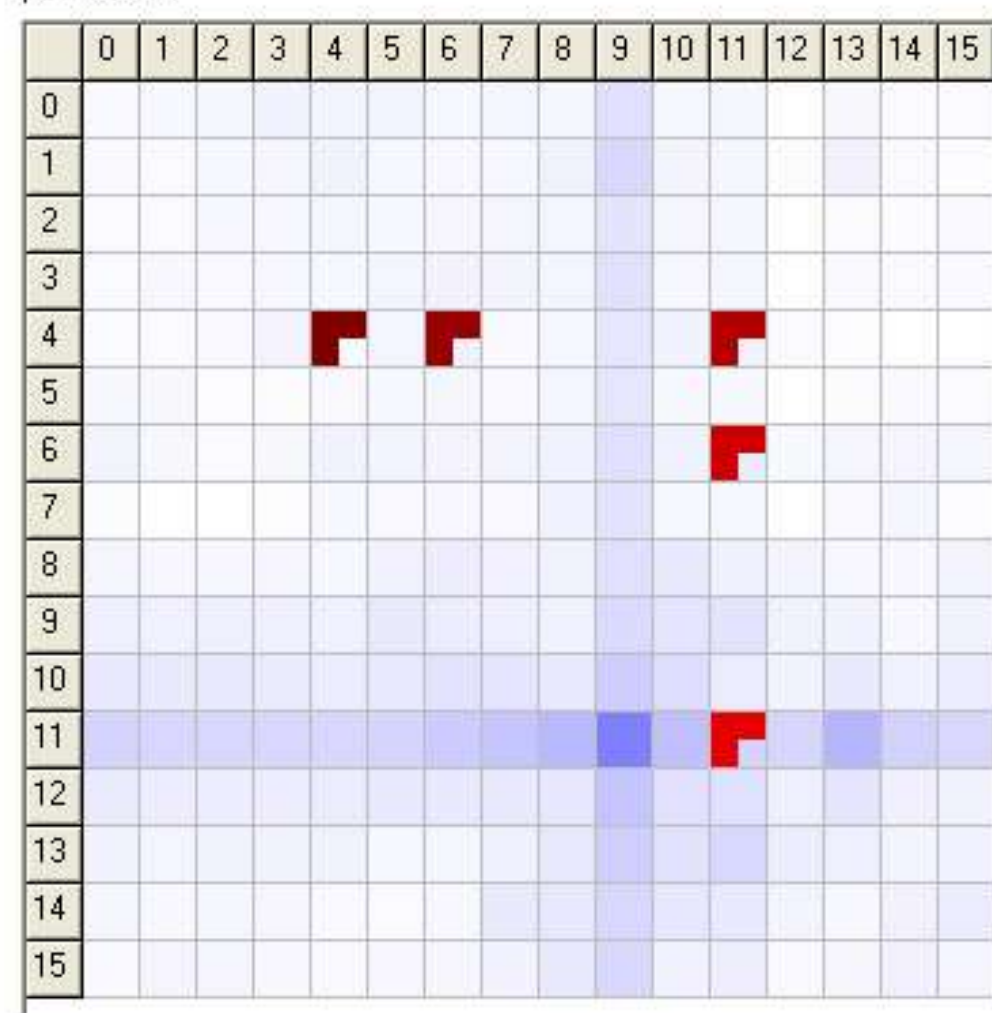

value distribution

| | |
|---|---|
| (>= 2.40) | - |
| ( 2.20) | - |
| ( 2.00) | 1 |
| ( 1.80) | - |
| ( 1.60) | - |
| ( 1.40) | - |
| ( 1.20) | - |
| ( 1.00) | - |
| ( 0.80) | - |
| ( 0.60) | - |
| ( 0.40) | 2 |
| ( 0.20) | 1 |
| ( 0.00) | 2 |
| ( -0.20) | 2 |
| ( -0.40) | 2 |
| ( -0.60) | 13 |
| ( -0.80) | 4 |
| ( -1.00) | 15 |
| ( -1.20) | 31 |
| ( -1.40) | 52 |
| ( -1.60) | 106 |
| (< -1.60) | 25 |

valuation by

```
[ 0] ( 1, 8)
[ 1] ( 4, 4)
[ 2] ( 8, 4)
[ 3] (10, 6)
[ 4] (10, 8)
```

sequence

```
[ 0] ( 1, 8)
[ 1] ( 4, 4)
[ 2] ( 8, 4)
[ 3] (10, 6)
[ 4] (10, 8)
```

top values 1. - 5.

```
( 9, 9):  0.1921
(11, 9): -0.7217
( 9,10): -0.7587
( 8,10): -0.7988
( 8, 9): -0.8500
```

positions

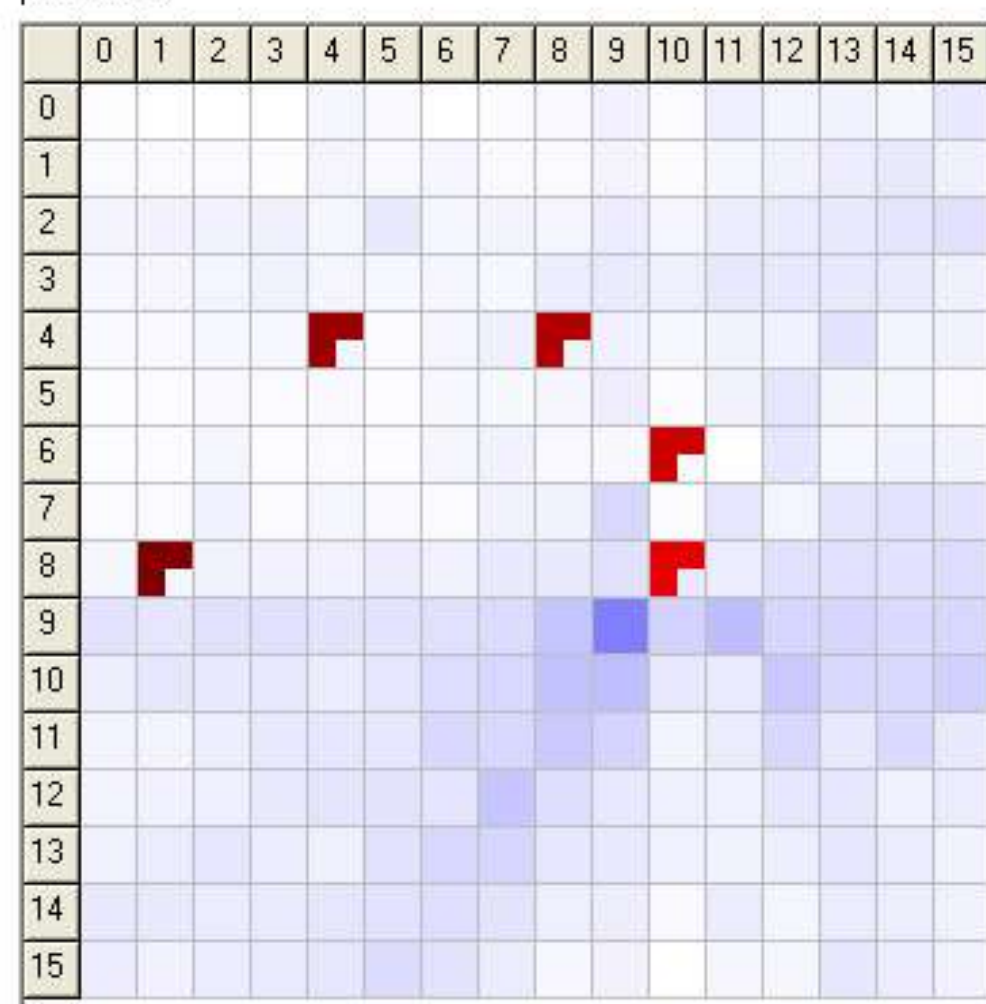

value distribution

| | |
|---|---|
| (>= 2.40) | - |
| ( 2.20) | - |
| ( 2.00) | - |
| ( 1.80) | - |
| ( 1.60) | - |
| ( 1.40) | - |
| ( 1.20) | - |
| ( 1.00) | - |
| ( 0.80) | - |
| ( 0.60) | - |
| ( 0.40) | - |
| ( 0.20) | - |
| ( 0.00) | 1 |
| ( -0.20) | - |
| ( -0.40) | - |
| ( -0.60) | - |
| ( -0.80) | 3 |
| ( -1.00) | 4 |
| ( -1.20) | 19 |
| ( -1.40) | 61 |
| ( -1.60) | 120 |
| (< -1.60) | 48 |

valuation by

```
[ 0] (10, 8)
[ 1] (11,10)
[ 2] ( 4, 9)
[ 3] ( 7, 8)
[ 4] ( 6,10)
```

sequence

```
[ 0] (10, 8)
[ 1] (11,10)
[ 2] ( 4, 9)
[ 3] ( 7, 8)
[ 4] ( 6,10)
```

top values 1. - 5.

```
( 6, 9): -0.5213
( 2, 4): -0.6569
(13, 9): -0.7449
(12, 9): -0.7782
( 0, 9): -0.7940
```

positions

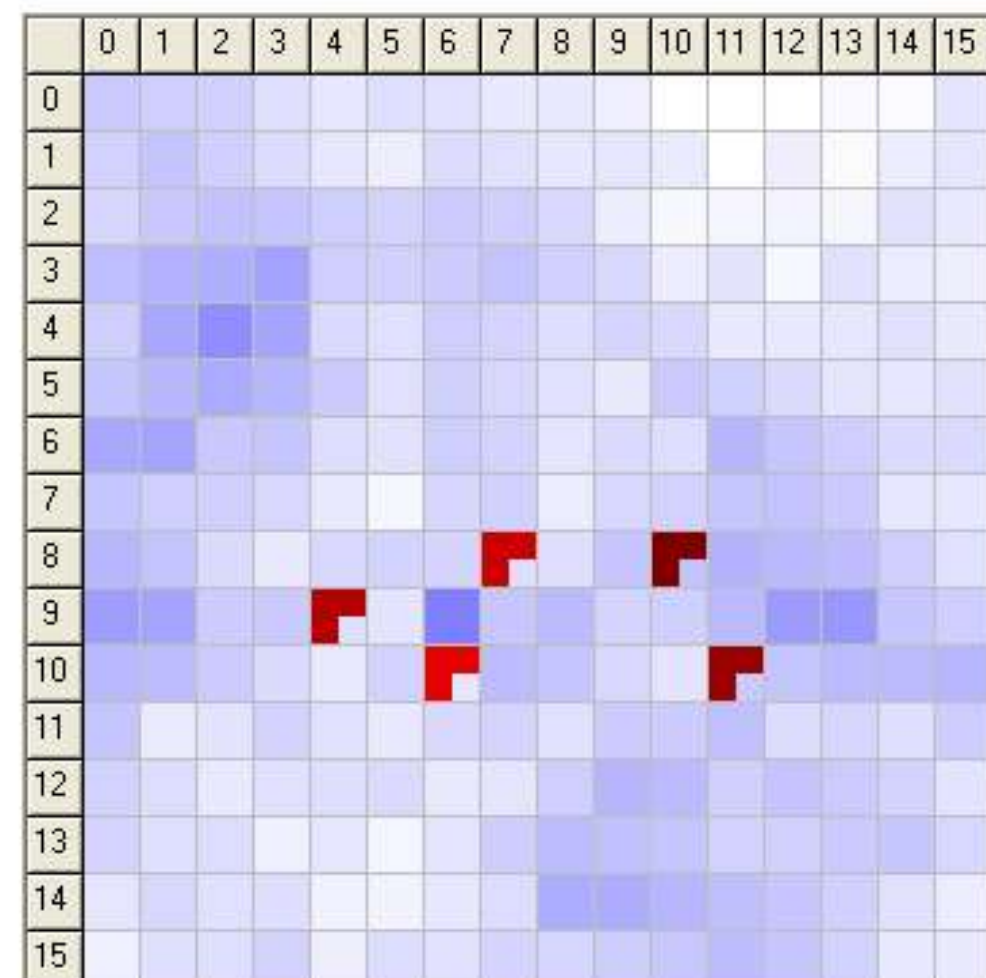

value distribution

| | |
|---|---|
| (>= 2.40) | - |
| ( 2.20) | - |
| ( 2.00) | - |
| ( 1.80) | - |
| ( 1.60) | - |
| ( 1.40) | - |
| ( 1.20) | - |
| ( 1.00) | - |
| ( 0.80) | - |
| ( 0.60) | - |
| ( 0.40) | - |
| ( 0.20) | - |
| ( 0.00) | - |
| ( -0.20) | - |
| ( -0.40) | - |
| ( -0.60) | 1 |
| ( -0.80) | 4 |
| ( -1.00) | 12 |
| ( -1.20) | 49 |
| ( -1.40) | 103 |
| ( -1.60) | 75 |
| (< -1.60) | 12 |

Now, the diagrams contain the sequence, which is the base of the valuation, the sequence, which is the base of continuations (both are equal in these examples), with position colors from dark red to light red, and continuations (positions with highest value) marked up by "*", with position colors from light green to dark green.

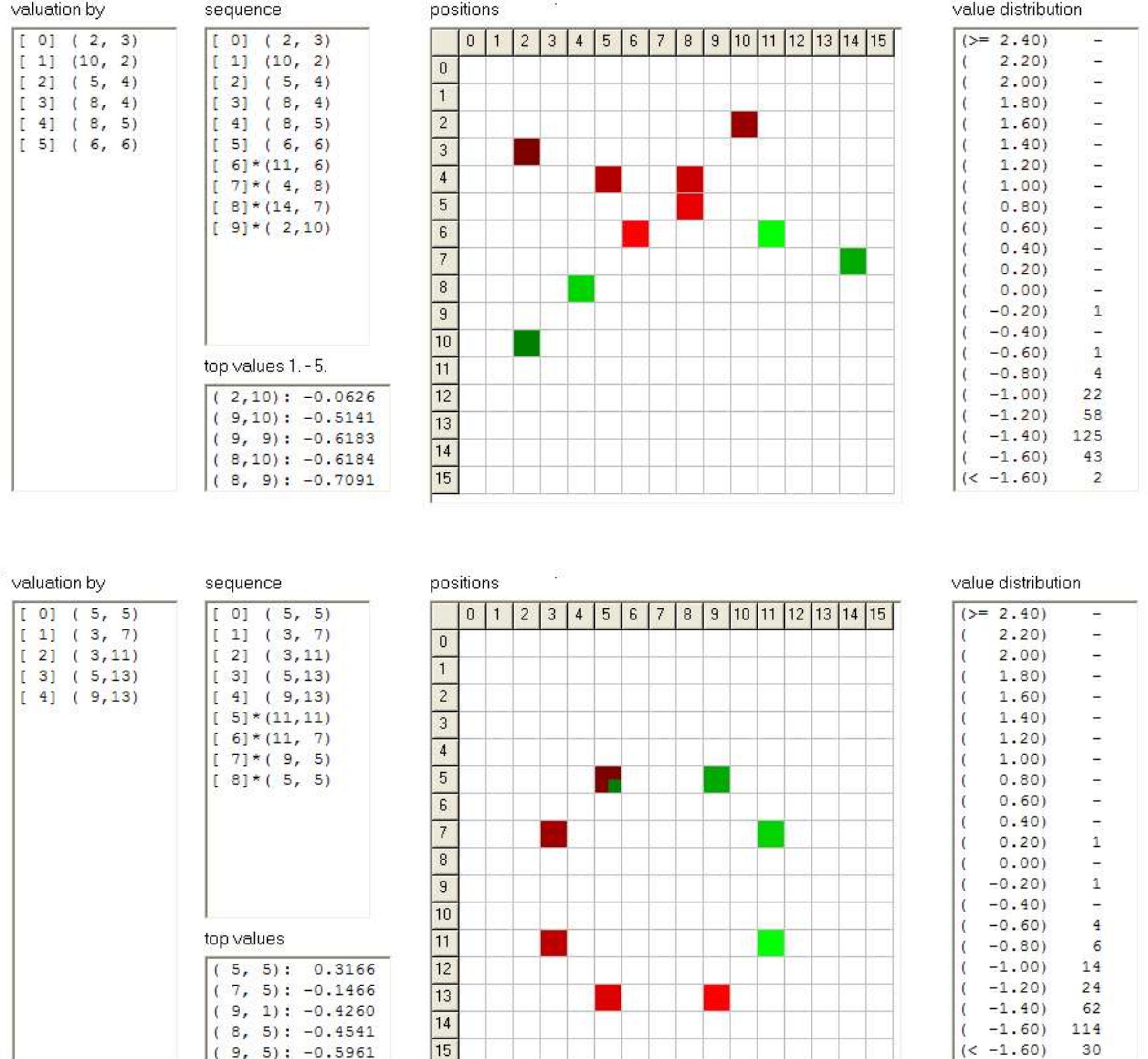

## 3. The Valuation Function

I need to valuate a position sequence c = $(c_0,...,c_{n-1})$. The positions are fields on the chess board, so they are complex numbers, field A1 = complex number (0,0), A2 = (0,1) ... B1 = (1,0) ... C7 = (2,6) ... . I need many different features of position sequences, based upon real valued operators on sequences of complex numbers. I will describe families of transformation steps. The real valued operators are combinations of selected transformations steps. These real valued operators lead to features by decomposition of the real axis into intervals. Each interval yields its contribution to the valuation function.

### 3.1. Example

I choose an operator φ, transforming sequences of complex numbers d = $(d_0, d_1, d_2)$ of length 3 to real values by

$$Op(d) := \text{real part } ((d_2 - d_1) / (d_1 - d_0))$$

The sequence c = $c_t$ of length n yields n-2 real values by its partial sequences of length 3 starting at t = 0, ... , n-3. On the other hand, I have a general sequence g, perhaps a random sequence of length 1000 (I used another “general sequence“, but this is not important). I want to compare the value distributions of all values of partial sequences of length 3. For this purpose, the axis of real numbers is partitioned into k disjoint intervals (k = 5, 6, ... or larger), by putting (nearly) the same number of values of the general sequence in each interval, i.e. a part of 1/k of all values per interval.

Now I build a special sequence s, see the examples above. This is the base of the valuations:

For a real number r there is an interval I in this partition, which contains r. The value of r compares the probabilities of the occurrence of the real values of the transformed sequences s and g in I.

$$V(r) := V(I) := \log (p(s \mid I) / p(g \mid I))$$

The value V(I) reflects whether an interval I is more preferred by values of sequence s or of sequence g. I have to avoid extreme values. Therefore I choose a global constant $\varepsilon > 0$, and before calculating V, I replace all probabilities < $\varepsilon$ by the value $\varepsilon$ itself. The choice of $\varepsilon$ is important for absolute values, but not for size relations between different valuation values.

### 3.2. Transformation of Position Sequences

#### *3.2.1. Convolution*

It is sufficient to use the following kind of convolutions: for an integer l > 0 I define the

convolution C by $C(c)_t := c_t + c_{t-1} + ... + c_{t-l+1}$

This operation replaces single numbers by "mean values" of consecutive numbers.

#### *3.2.2. Invariant Features*

For an integer l > 0

difference D: $D(c)_t := c_t - c_{t-1}$

quotient Q: $Q(c)_t := c_t / c_{t-1}$

(:= (0,0), if $c_{t-1} = 0$; but this is not important in our examples)

symm. quot. SQ: $SQ(c)_t := c_t / c_{t-1} + c_{t-1} / c_t$

Application of D leads to translation invariance. Application of D, then Q, leads to translation rotation stretching invariance, because such a transformation is invariant in the following sense: let a,b be complex numbers, a not 0. I define a new sequence d by

$d_t := a * c_t + b$

Then there is

$(Q \circ D)(c) = (Q \circ D)(d)$

*3.2.3. Transition to Real Numbers*

$(R \circ c)_t := R(c_t)$

R is one of the following operators:

R(z) := real part of z
R(z) := imaginary part of z
R(z) := modulus of z
R(z) := angle of z

**3.3. Combination of Transformations**

An operator Op consists of a convolution C, difference D, a transformations T = combination of D, Q and SQ, and a transition to real numbers R.

$Op = R \circ T \circ D \circ C$

The selection of suitable operators must be independent of special sequences or considerations upon special sequences. It may depend on the expected length of the sequences. The selection of operators, which is used in the present case, follows these lines: I use equal numbers of convolutions of length 1, 2, 3, 4. The part T of Op is one of the transformations I (identity), D, Q, D ° Q, Q ° Q, SQ, D ° SQ, Q ° SQ. Operators with T = I, T = D may have less invariances, Operators with SQ have more invariances.

**3.4. Valuation of Prolongations**

I apply Op to the last partial sequence end(prolongation of c) of a given sequence c, which contains the new last point of c (the continuation) and which is long enough, so that Op produces a result. The value of the prolongation is the mean value of all values V(Op(end(prolongation of c))), where V is the valuation derived from Op by the same or another sequence, see 3.1. This is the valuation of each prolongation of the sequence c.

**3.5. Valuation of the Similarity between Sequences**

I get a valuation of an abitrary sequences d, if I take the mean value over all values V(Op(part(d))), where Op runs through all operators, V is the valuation derived from Op, and part(d) runs through all partial sequences of d with the suitable lenght to apply Op. This mean value measures the similarity between the sequence c and the sequence d.

## 4. Results and Further Applications

### 4.1. The Effect of Feature Selection

In [5] I compared the quality of different methods of feature selection. The first set of features consists of convolutions only; the second also uses differences and the third one uses quotients of differences instead of differences. I calculated the rank of the prescribed continuations of 15 sequences of length 6 to sequences of length from 7 to 12, i.e. 90 continuations. For a single continuation, there are 144 continuations available, so in the random case the mean rank of a selected continuation will be 72.5. With the first set of features the mean rank is approximately 40, with the second set approximately 10 and the third set approximately 3. With the more complete feature selection described in part 3, the mean rank is approximately 1.5. These results are stable although the selection process contains random steps. The continuations of the sample sequences used in [5] are not at all unique because this allows for a more subtle observations about different methods, which I wanted to compare. A mean rank of 1.5 is an acceptable result.

### 4.2. Other Valuation Functions

I used quotients log (p(s | I), p(g | I)) as numerical contributions of the features, see 3.4. In [5] I compared this choice with others. The simplest choice is the value 1, if p > 0, and 0, if p = 0. I have observed that for strongly regular sequences all choices lead to nearly equal results. Therefore, the log function seems to be too complicated. But for long sequences and for less regular sequences it is best suited, especially as a memory for random sequences. For random sequences, the features tend to be independent, so the sum of log (p(s | I), p(g | I)) has a real meaning. It is the log of the quotient of the probabilities of special events with all these features and the same of general events. I expect, that in a more general context the behaviour near random cases is more important, than the behaviour in  strongly regular cases.

### 4.3. Similarity of Sequences

The valuation of possible continuations valuates the similarity of different sequences, too. Another application is as follows. I set up a position sequence. Here it resembles staircase. I calculate the corresponding valuation. Then I set up a short sequence. I continue this sequence by these positions, which are preferred by the given valuation gained in the first staircase. The continued sequence should resemble this first set of stairs.

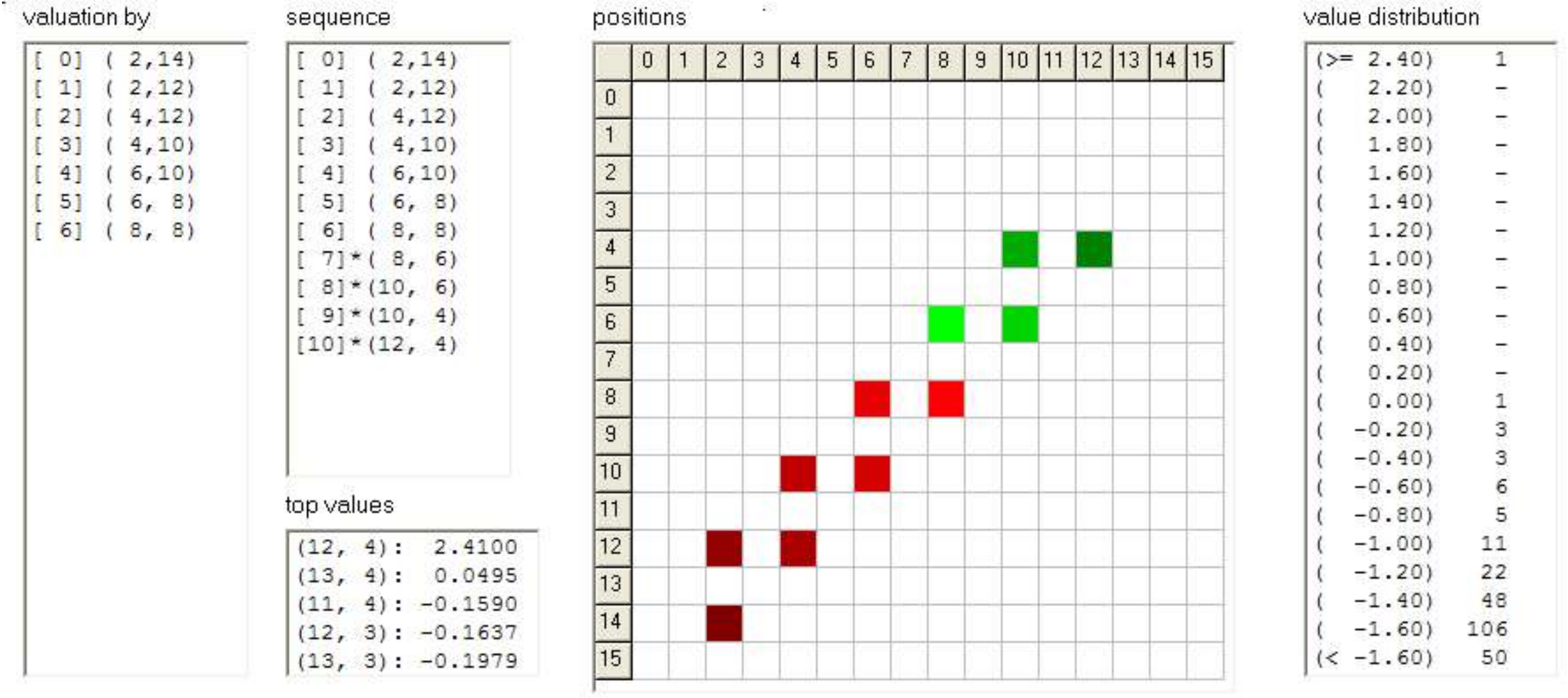

valuation by

```
[ 0]  ( 2,14)
[ 1]  ( 2,12)
[ 2]  ( 4,12)
[ 3]  ( 4,10)
[ 4]  ( 6,10)
[ 5]  ( 6, 8)
[ 6]  ( 8, 8)
```

sequence

```
[ 0]  ( 4, 3)
[ 1]  ( 3, 5)
[ 2]  ( 6, 5)
[ 3]  ( 5, 7)
[ 4]*( 8, 7)
[ 5]*( 7, 9)
[ 6]*(10, 9)
[ 7]*( 9,11)
[ 8]*(12,11)
[ 9]*(11,13)
```

top values

```
(11,13):   0.6455
(12,14):   0.4946
(12,13): -0.2802
(13,15): -0.3859
(12,12): -0.4569
```

positions

| | 0 | 1 | 2 | 3 | 4 | 5 | 6 | 7 | 8 | 9 | 10 | 11 | 12 | 13 | 14 | 15 |
|---|---|---|---|---|---|---|---|---|---|---|---|---|---|---|---|---|
| 0 | | | | | | | | | | | | | | | | |
| 1 | | | | | | | | | | | | | | | | |
| 2 | | | | | | | | | | | | | | | | |
| 3 | | | | | | | | | | | | | | | | |
| 4 | | | | | | | | | | | | | | | | |
| 5 | | | | | | | | | | | | | | | | |
| 6 | | | | | | | | | | | | | | | | |
| 7 | | | | | | | | | | | | | | | | |
| 8 | | | | | | | | | | | | | | | | |
| 9 | | | | | | | | | | | | | | | | |
| 10 | | | | | | | | | | | | | | | | |
| 11 | | | | | | | | | | | | | | | | |
| 12 | | | | | | | | | | | | | | | | |
| 13 | | | | | | | | | | | | | | | | |
| 14 | | | | | | | | | | | | | | | | |
| 15 | | | | | | | | | | | | | | | | |

value distribution

```
(>= 2.40)    -
(   2.20)    -
(   2.00)    -
(   1.80)    -
(   1.60)    -
(   1.40)    -
(   1.20)    -
(   1.00)    -
(   0.80)    -
(   0.60)    1
(   0.40)    1
(   0.20)    -
(   0.00)    -
(  -0.20)    -
(  -0.40)    2
(  -0.60)    1
(  -0.80)    5
(  -1.00)    9
(  -1.20)   27
(  -1.40)   58
(  -1.60)  126
(< -1.60)   26
```

valuation by

```
[ 0]  ( 2,14)
[ 1]  ( 2,12)
[ 2]  ( 4,12)
[ 3]  ( 4,10)
[ 4]  ( 6,10)
[ 5]  ( 6, 8)
[ 6]  ( 8, 8)
```

sequence

```
[ 0]  (13,14)
[ 1]  (12,13)
[ 2]  (12,11)
[ 3]  (11,10)
[ 4]*(11, 8)
[ 5]*(10, 7)
[ 6]*(10, 5)
[ 7]*( 9, 4)
[ 8]*( 9, 2)
```

top values

```
( 9, 2):   0.5676
(10, 2): -0.2704
( 9, 3): -0.3432
(10, 3): -0.3606
(11, 2): -0.4329
```

positions

| | 0 | 1 | 2 | 3 | 4 | 5 | 6 | 7 | 8 | 9 | 10 | 11 | 12 | 13 | 14 | 15 |
|---|---|---|---|---|---|---|---|---|---|---|---|---|---|---|---|---|
| 0 | | | | | | | | | | | | | | | | |
| 1 | | | | | | | | | | | | | | | | |
| 2 | | | | | | | | | | | | | | | | |
| 3 | | | | | | | | | | | | | | | | |
| 4 | | | | | | | | | | | | | | | | |
| 5 | | | | | | | | | | | | | | | | |
| 6 | | | | | | | | | | | | | | | | |
| 7 | | | | | | | | | | | | | | | | |
| 8 | | | | | | | | | | | | | | | | |
| 9 | | | | | | | | | | | | | | | | |
| 10 | | | | | | | | | | | | | | | | |
| 11 | | | | | | | | | | | | | | | | |
| 12 | | | | | | | | | | | | | | | | |
| 13 | | | | | | | | | | | | | | | | |
| 14 | | | | | | | | | | | | | | | | |
| 15 | | | | | | | | | | | | | | | | |

value distribution

```
(>= 2.40)    -
(   2.20)    -
(   2.00)    -
(   1.80)    -
(   1.60)    -
(   1.40)    -
(   1.20)    -
(   1.00)    -
(   0.80)    -
(   0.60)    -
(   0.40)    1
(   0.20)    -
(   0.00)    -
(  -0.20)    -
(  -0.40)    3
(  -0.60)    2
(  -0.80)    4
(  -1.00)   15
(  -1.20)   23
(  -1.40)   69
(  -1.60)  117
(< -1.60)   22
```

### 4.4. Rules and Regularity

I create a sequence of moves by rooks by a random rook move generator. I investigate the valuation function of this sequence. The question is whether the principle “move by a rook” will be recognised. This is the case when the following applies: If a sequence of moves by rooks is given and the continuation possibilities are sorted according to their valuation, then the allowed moves by a rook (30 moves) must come out on top. The result is: I need a large number of rook moves as base of the valuation. In each case, not to move is valuated as a rook move, too. The method "EE" is not appropriate to find single rules.

The examples contain 20 and 200 rook moves.

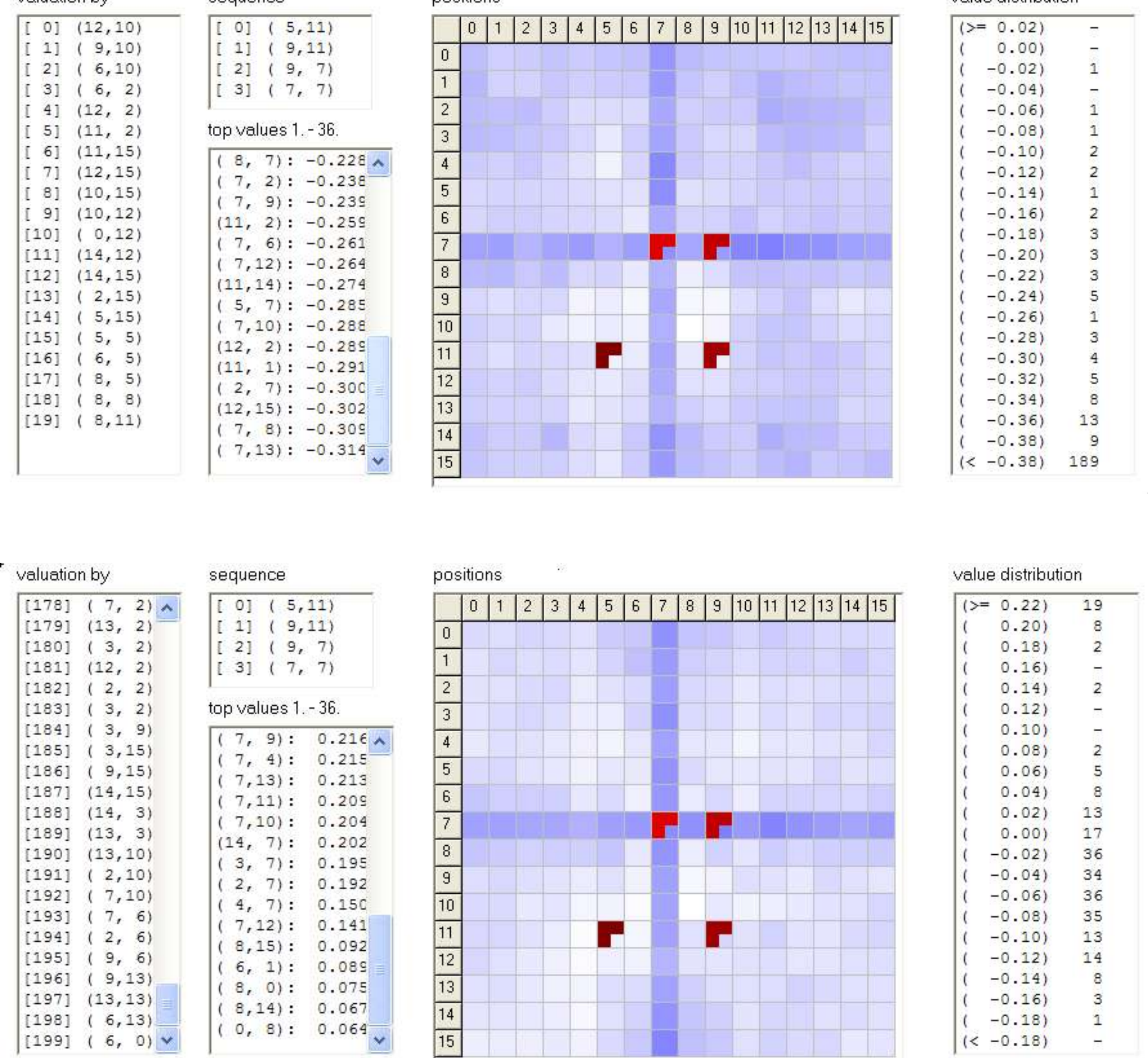

## 4.5. Memory vs. Regularity

I also examined how the individual positions themselves are valuated as a continuation of an initial part of the sequence. The valuation function serves as “memory” for the sequence if it achieves the following: If one stipulates an initial section of this sequence (for example, its first 5 positions), the continuations with the highest values are exactly the sequence positions at that place. The valuation function can therefore be used to reconstruct a sequence from an initial section. The valuation function still proves to be reliable memory in sequences with a length of 50. With a length of 100 one to two deviations can be expected (results in [5]). However, in short sequences a conflict can arise between memory and the recognition of regularities. For that reason one cannot expect a perfect memory. I would like to demonstrate this with an example.

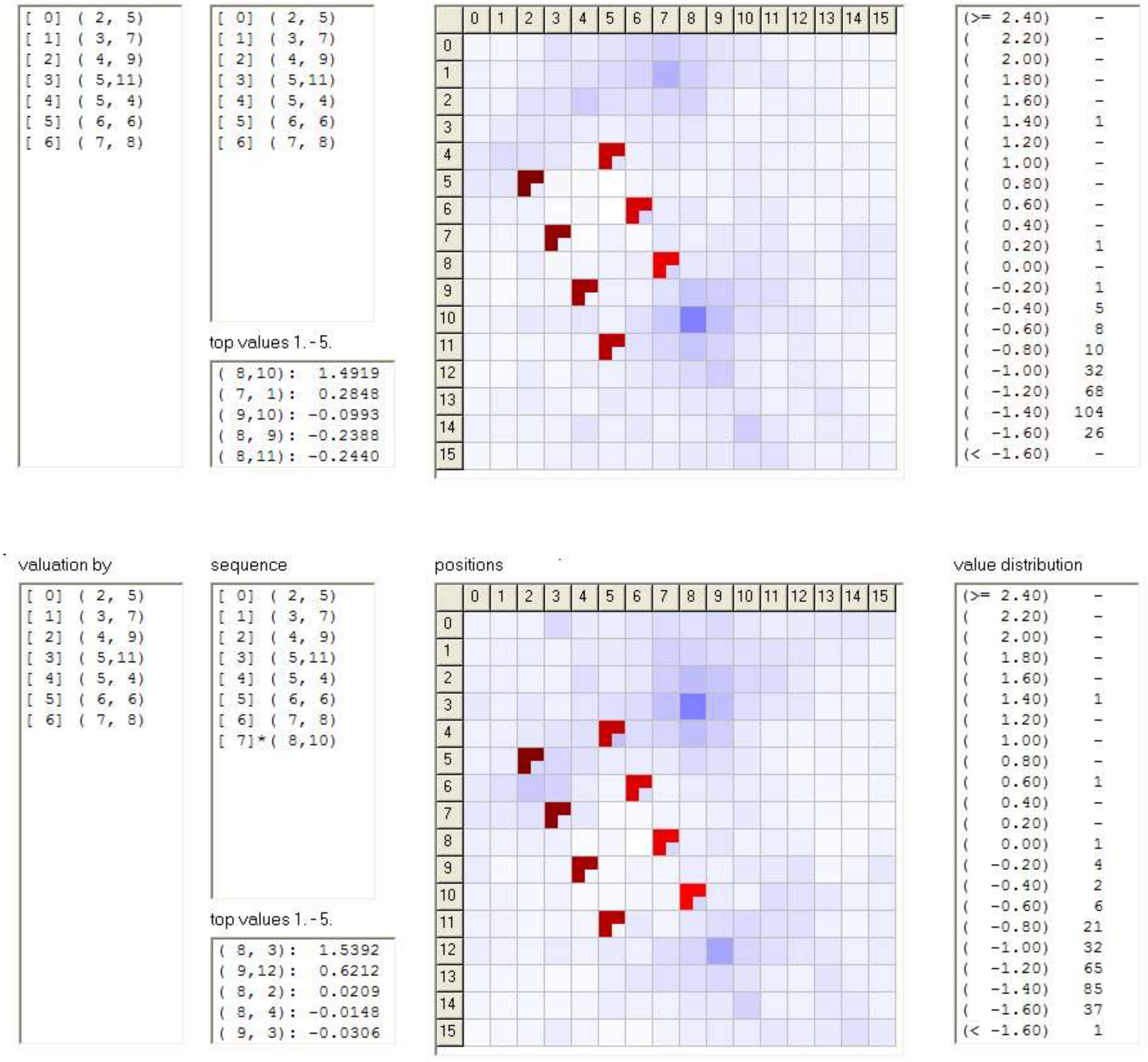

## 5. Prospects

There are many more options in the choice of features. For instance, operators can be used that are invariant under any affine transformations, not only under translation, rotation and stretching. This is an entirely appropriate invariance if one considers sequences of points in space that are projected onto a surface. I have experimented with this but could not ascertain any improvements. I believe the reason is that such invariants require sequences of length at least 4 while here the model used can manage sequences of length 3 and for translation invariance even of length 2. In a certain sense, the operator which measures the distance between two consecutive positions is ideal for short sequences; it requires only length 2 and still has relatively many invariances. For longer sequences, on the other hand, longer operators definitely may be used.

It is well known, that our visual system is able to produce invariant features, see [6], where the pinwheel structure of neurons leeds to locally rotation invariant results.

Impressive experiments have been performed in which it has been demonstrated that pigeons can employ invariants, and even ignore them when necessary in order to distinguish between sample objects (see [3]).

There are numerous methods for labelling images by means of invariant features (see [7], [8]), where the object of investigation is invariants of sets of points. In my opinion, invariants of point sequences are much more accessible. For that reason it is desirable to examine the role they play in cognitive processes in more detail. A close relative to a point sequence is a trajectory (curve), which has also been thoroughly examined (a patented procedure for handwriting recognition in [1]). Discrete sequences of points are used in each approach to the numerical treatment of trajectories. Thus they are also in principle accessible to my methods.

If it is assumed, that it is impossible to cover uniformly the total range of continuing sequences, then it is also impossible to obtain reliable quality standards for various approaches to solutions. In my opinion, this is not a fault but rather inherent to the problem. There is no one single method for different persons with which each individual must achieve the same result. Of course, there are examples in which that is the case but there are counter-examples, as well, in which the choice of continuation is disputable. I consider it appropriate, although requiring much familiarisation, if this is reflected in the model chosen for the valuation function (distribution of features). I do not expect that there is “the” concept of regularity, not even in the special case of the continuation problem. Rather, we should follow the suggestion of A. Turing and ask: If an unknown system, faced with sequences of positions on a chess board, can respond and give its preferred continuation, is it then possible to determine from the answers whether the system is a person or a computer program?